\documentclass{article}

\usepackage[final, nonatbib]{neurips_2024}

\usepackage[utf8]{inputenc} 
\usepackage[T1]{fontenc}    
\usepackage{hyperref}       
\usepackage{url}            
\usepackage{booktabs}       
\usepackage{amsfonts}       
\usepackage{nicefrac}       
\usepackage{microtype}      
\usepackage{xcolor}         

\makeatletter
\newcommand*{\rom}[1]{\expandafter\@slowromancap\romannumeral #1@}
\makeatother
\usepackage{enumitem}
\usepackage{comment}

\usepackage{graphicx}
\usepackage{amsmath} 
\usepackage{float}
\usepackage{wrapfig}
\usepackage[ruled]{algorithm2e}
\usepackage{multirow,bm}
\usepackage{amsthm}

\theoremstyle{plain}
\newtheorem{theorem}{Theorem}[section]

\theoremstyle{definition}

\theoremstyle{remark}

\DeclareMathOperator*{\argminA}{arg\,min}

\newif\ifmodify 
\modifytrue 
\ifmodify

\else

\fi

\title{
Search for Efficient Large Language Models
}

\author{
  Xuan Shen$^1$,
  Pu Zhao$^1$,
  Yifan Gong$^1$, Zhenglun Kong$^2$, Zheng Zhan$^1$, \\
  \textbf{Yushu Wu$^1$, Ming Lin$^3$, Chao Wu$^1$, Xue Lin$^1$, Yanzhi Wang$^1$}\\
  $^1$Northeastern University, 
  $^2$Harvard University,
  $^3$Oracle \\
  \texttt{\{shen.xu, yanz.wang\}@northeastern.edu} \\
}

\begin{document}

\maketitle

\begin{abstract}


Large Language Models (LLMs) have long held sway in the realms of artificial intelligence research.
Numerous efficient techniques, including weight pruning, quantization, and distillation, have been embraced to compress LLMs, targeting memory reduction and inference acceleration, which underscore the redundancy  in LLMs.
However, most model compression techniques concentrate on weight optimization, overlooking the exploration of optimal architectures.
Besides, traditional architecture search methods, limited by the elevated complexity with extensive parameters, struggle to demonstrate their effectiveness on LLMs.
In this paper, we propose a training-free architecture search framework to identify optimal subnets that preserve the fundamental strengths of the original LLMs while achieving inference acceleration.
Furthermore, after generating subnets that inherit specific weights from the original LLMs, we introduce a reformation algorithm that utilizes the omitted weights to rectify the inherited weights  with a small amount of calibration data.
Compared with SOTA training-free structured pruning works that can generate smaller networks, our method demonstrates superior performance across standard benchmarks.
Furthermore, our generated subnets can directly reduce the usage of GPU memory and achieve inference acceleration.
Code: \url{https://github.com/shawnricecake/search-llm}

\end{abstract}


\section{Introduction}

Large Language Models (LLMs)~\cite{touvron2023llama, zhang2022opt, scao2022bloom, brown2020language, radford2019language, gpt3} are renowned for their exceptional performance across various domains of artificial intelligence research.
There is a growing demand for constructing LLMs for extensive applications across a multitude of popular platforms.
However, the computational and storage costs have restricted  LLMs from deployment on various devices for wide applications.
Take the GPT-3 model as an example, with its 175 billion parameters~\cite{gpt3}, it requires more than 326GB of memory in FP16 format. This exceeds the memory capabilities of even the most sophisticated GPUs, far surpassing available memory on resource-constrained devices.
To address these challenges, a variety of compression techniques focusing on weight optimization have been developed, including weight pruning~\cite{frantar-sparsegpt, ma2023llmpruner, ma2023llmpruner, ashkboos2024slicegpt, an2023flap,zheng2024exploring, shen2024edgeqat,yang2023pruning,zhang2022advancing,rtseg,zhao2024pruningfoundationmodelshigh}, quantization~\cite{frantar-gptq, xiao2023smoothquant, Shen_2024_Agile}, and knowledge distillation~\cite{sun2019patient, sun-etal-2020-contrastive, pan-etal-2021-meta}. The extensive research in the compression direction indicates the substantial redundancy within LLMs.


Besides optimizing model weights, improving the model architecture is another crucial direction in achieving both high efficiency and superior performance. Numerous works~\cite{pmlr-v97-tan19a, Shen_2023_CVPR, ming_zennas_iccv2021, gong2022nasvit, su2021vision, Ma_2023_CVPR, autoFormer, NEURIPS2023_081b0806,zhan2021achieving,wu2022compiler, yang2023fast, yang2023late,dong2024hotbev,li2022pruning} have studied the Neural Architecture Search (NAS) problem for representative model designs such as Convolutional Neural Networks (CNNs) and Vision Transformers (ViTs). However, the realm of architecture search for LLMs remains unexplored. Though enjoying the potential benefits of discovering highly efficient and well-performing LLM architectures compared with manual designs, searching with traditional NAS methods for LLMs faces significant challenges due to the immense complexity and extensive model size. Furthermore, the convergence of a randomly initialized architecture to the searched optimal state takes substantial training efforts and resources, intensifying the challenges of searching for efficient LLM architectures.  

To tackle the challenges, we propose a training-free architecture search framework that discovers efficient LLM architectures within the original well-trained LLMs. Specifically, to reduce the search cost, we first identify an appropriate initial architecture by computing the importance of weights. Subsequently,  an evolution-based algorithm is applied to globally search an efficient subnet starting from the initial subnet. In each generation, mutation and crossover are adopted to generate candidate architectures within the search space. The candidates are evaluated efficiently with a small amount of training samples to assess their effectiveness and select for the next generation. As we start our search from a well-trained LLM instead of randomly initialized models, we propose a mask mutation algorithm to identify the detailed channel indices, rather than just the number of channels in the mutation of traditional NAS~\cite{autoFormer, Shen_2023_CVPR, ming_zennas_iccv2021, gong2022nasvit,wu2022compiler}. 
After a few generations with the identified promising LLM architecture, we adopt the reformation algorithm based on the alternating direction method of multipliers (ADMM) \cite{parikh2014proximal,boyd2011distributed,10.1145/3240508.3240639,gong2020privacy} to rectify weights in inherited efficient LLM architecture with omitted weights (i.e., non-inherited weights) by leveraging only 128 calibration samples.

\begin{figure*}[t]
  \centering
  \includegraphics[width=1.0\textwidth]{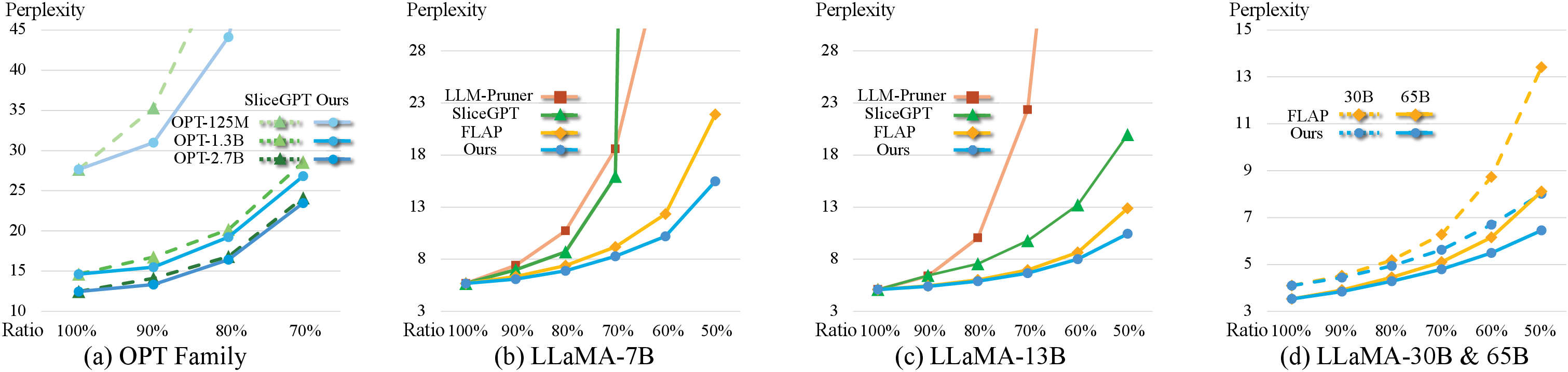}
  \vspace{-0.7cm}
  \caption{
  Experiment results of perplexity $\downarrow$ on WikiText2 dataset with 2048 sequence length.
  }
  \label{fig:sota-visual}
  \vspace{-0.4cm}
\end{figure*}

As shown in Figure~\ref{fig:sota-visual}, our extensive experiments demonstrate that our method can achieve superior performance than SOTA structured pruning baselines in terms of perplexity and zero-shot accuracy on multiple datasets across various LLM families and model sizes.
Particularly, as in Figure~\ref{fig:sota-visual}~(a), only SliceGPT and our method support the OPT model family, and our method outperforms SliceGPT. Additionally, with a 60\% inheriting ratio for the LLaMA-7B model on the WikiText2 dataset, our method achieves the best performance with a perplexity of 10.21, compared to 38.27 by LLM-Pruner and 279.52 by SliceGPT, as illustrated in Figure~\ref{fig:sota-visual} (b).
Furthermore, when scaling to LLaMA-13B, both SliceGPT and LLM-Pruner fail, as in Figure~\ref{fig:sota-visual} (c). Lastly, as in Figure~\ref{fig:sota-visual}~(d), only FLAP and our method support the LLaMA-30B and 65B models, and our method achieves better performance than FLAP.
Besides, our implementations on GPUs demonstrate significant memory reduction and inference acceleration.
Meanwhile, our approach eliminates the retraining process, relying solely on forward pass for both searching and reformation processes, which maintains a low memory overhead.
Our contributions are summarized below,

\begin{itemize}[label={}, leftmargin=*]

\item \textbf{1.} We propose a training-free search framework to identify subnets within LLMs, featuring an importance-aware initialization that significantly reduces the time cost of searching, and an evolution architecture search with special mask mutation and efficient candidate evaluation. 

\item \textbf{2.} We propose a reformation algorithm that reconstructs weights by calibrating with only 128 training samples, thereby enhancing the effectiveness of the subnets.

\item \textbf{3.} Experiments indicate that the subnets generated by our method outperform SOTA structured pruning works in terms of perplexity and accuracy on multiple datasets across various LLM families and sizes. The searched subnets can effectively reduce GPU memory and accelerate inference.


\end{itemize}

\section{Related Work}

\subsection{Compression of LLMs}

Various compression techniques have been developed to reduce the model size or inference cost of LLMs, including model pruning, quantization, and distillation.
Among these, quantization and structured pruning methods are  prevalent due to their efficacy in inference acceleration while preserving task performance.
Quantization approaches, as explored in works~\cite{Shen_2024_Agile, xiao2023smoothquant, frantar-gptq}, compress models by converting weights to lower bit representations.
Besides, structured pruning techniques, including the works~\cite{ma2023llmpruner, ashkboos2024slicegpt, an2023flap, sun2023wanda}, remove redundant weights in a structured manner to reduce the total weight count.
Specifically, LLM-Pruner~\cite{ma2023llmpruner} eliminates non-critical coupled structures based on gradient information, while SliceGPT~\cite{ashkboos2024slicegpt} substitutes each weight matrix with a smaller, dense matrix and reduces the embedding dimension of the network.
FLAP~\cite{an2023flap} employs structured metrics to prune LLMs and globally optimizes the sparsity ratio with the output feature maps.
Despite the advancements, most pruning methods indiscriminately remove heads within the self-attention modules, leading to more significant performance loss due to  the inherent input-dependent nature of transformer architectures based on all heads.

\subsection{Search for Transformers}

NAS has emerged as a pivotal technique for identifying efficient architectures in CNNs (exemplified by EfficientNet~\cite{tan2019efficientnet}) and  transformer-based models (such as BERT~\cite{devlin-etal-2019-bert} and Vision Transformer~\cite{dosovitskiy2020vit}).
To mitigate the typical high training costs  associated with NAS, innovations such as one-shot and zero-shot NAS methods~\cite{gong2022nasvit, autoFormer, ming_zennas_iccv2021, Shen_2023_CVPR} have been developed, enhancing the efficiency of generating high-performance architectures.
In contrast to zero-shot NAS methods, which utilize accuracy predictors to derive optimal architectures, one-shot NAS methods streamline the process by pretraining a comprehensive supernet from which optimal subnets are subsequently selected.
Specifically, in the context of transformer-based models, the one-shot NAS approach, as implemented in AutoFormer~\cite{autoFormer}, involves multiple rounds of supernet training, strategically extending weights along certain dimensions to optimize performance.
NASViT~\cite{gong2022nasvit} leverages gradient information during supernet training to refine subnet selection and mitigate gradient conflicts, thereby enhancing the effectiveness of generated architectures.
The proven efficacy of one-shot NAS for transformer architectures provides a compelling rationale for its application to LLMs, considering that pretrained LLMs can function analogously as supernets.
This adaptation holds the potential to significantly advance the development and optimization of LLM architectures, motivating us to refine and enhance the capabilities of these complex models.

\begin{figure*}[t]
\vspace{-0.3cm}
  \centering
  \includegraphics[width=1.0\textwidth]{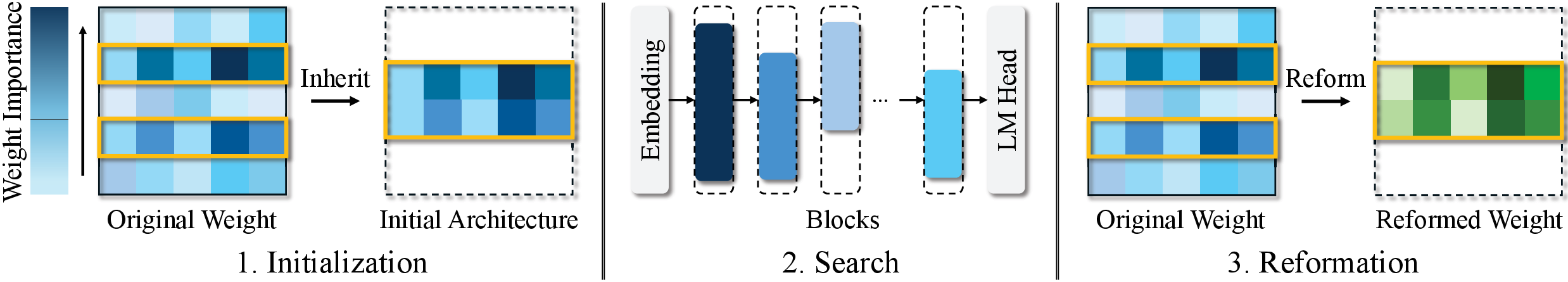}
  \vspace{-0.7cm}
  \caption{
  Framework Overview.
  }
  \label{fig:framework_overveiw}
  \vspace{-0.4cm}
\end{figure*}

\section{Methodology}

\subsection{Framework Overview}
We show the overview of our search framework in Figure~\ref{fig:framework_overveiw}. It comprises three key components: search initialization, search pipeline, and weight reformation. First, an initial efficient architecture is constructed layer by layer with a uniform inheriting ratio based on the weight importance. Subsequently, based on the initialization, we conduct a comprehensive search process for the globally efficient architecture with the evolution-based search method. Finally, a reformation method is introduced to enhance the performance of the resulting subnets in   LLMs without retraining.

\subsection{Search Initialization} \label{sec:intialization}

\textbf{Global search with uniform initialization.} 
Unlike prior efficient LLM research efforts   such as SparseGPT \cite{frantar-sparsegpt} with a uniform sparsity ratio across all layers, our method leverages a global search approach, such that different layers in our searched architecture may inherit different percentages of parameters (inheriting ratios) from the original LLM. To reduce the search cost and promote the search performance, we initialize our search with the same inheriting ratio for all layers.
Through our search process, we iteratively refine the architecture, yielding subnets with varied inheriting ratios across layers. We demonstrate the pivotal role of the initialized architecture in driving search efficiency and effectiveness in Figure \ref{fig:ablation_search_process} and Section \ref{sec:pipeline}.  

\begin{figure*}[t]
\vspace{-0.3cm}
  \centering
  \includegraphics[width=1.0\textwidth]{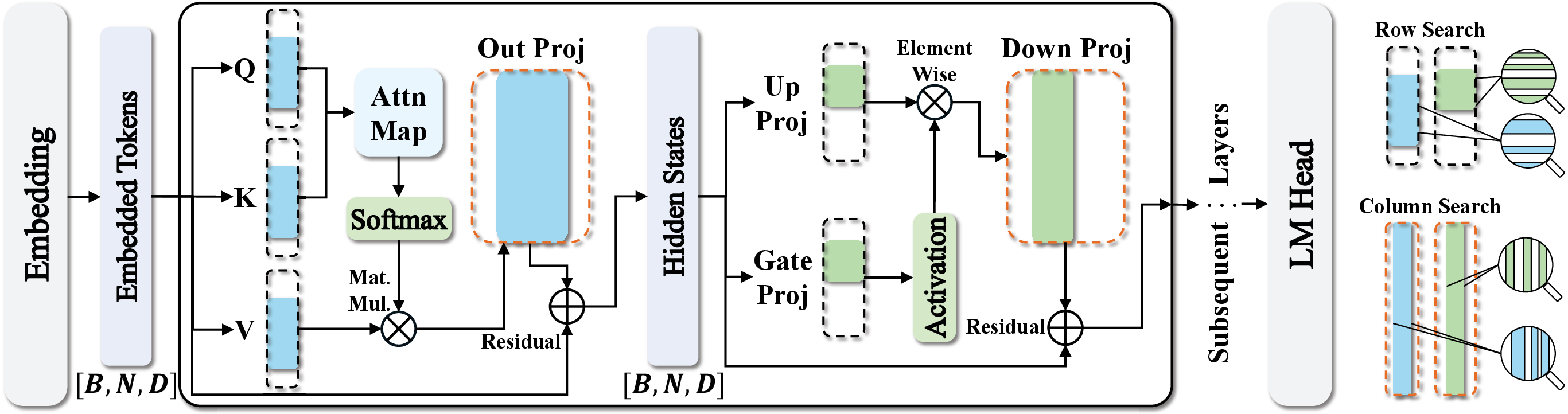}
  \vspace{-0.7cm}
  \caption{
  Visualization of the subnets generation for LLaMA family based on the selections masks $\mathbf{S}_{attn}$ for the self-attention module colored in blue and $\mathbf{S}_{mlp}$ for the MLP module colored in green.
  }
  \label{fig:search_overview}
  \vspace{-0.4cm}
\end{figure*}

\textbf{Structural subnets. }
To enable efficient inference,  we search structural subnets from the original LLMs, i.e., certain rows or columns in the original 2D weight matrix are inherited in our searched model. Take the LLaMA family as an example. In each attention block of LLaMA models, there are query, key, value, and output linear layers in the self-attention module with weights denoted by $\mathbf{W}_{Q}$, $\mathbf{W}_{K}$, $\mathbf{W}_{V}$, and $\mathbf{W}_{O}$, respectively, and other three linear layers $\mathbf{W}_{U}$, $\mathbf{W}_{G}$, and $\mathbf{W}_{D}$ in the MLP module.   To ensure the consistent hidden size in LLaMA models, based on the computation patterns in each block, we select rows in $\mathbf{W}_{Q}$, $\mathbf{W}_{K}$, $\mathbf{W}_{V}$,  $\mathbf{W}_{U}$, and $\mathbf{W}_{G}$, and columns in $\mathbf{W}_{O}$ and  $\mathbf{W}_{D}$. More details are presented in Figure \ref{fig:search_overview} and  Appendix \ref{app:sec:small_dense}. 


\textbf{Initialization based on importance score. }
We construct the initial building blocks by inheriting appropriate rows/columns from the original LLM layer. To determine which row/column to be inherited, we compute the importance score for each row and column as below, 
\begin{equation}
    [\Phi^{r}_{\mathbf W}]_i = \sum_{j} [\mathbf \Phi]_{i,j}, \  \ \ \
   [\Phi^{c}_{\mathbf W}]_j = \sum_{i} [\mathbf \Phi]_{i,j},  \  \ \ \
        [\mathbf \Phi]_{i,j} = \frac{  [\mathbf{W}]_{i,j}^2 }{   [( 2   \mathbf  X   \mathbf X^T)^{-1}]_{j,j}  },
\end{equation}
where $[\Phi^{r}_{\mathbf W}]_i$  represents  the row score for the  $i^{th}$ row of $\mathbf{W}$  and $[\Phi^{c}_{\mathbf W}]_j$ denotes the column score of  the  $j^{th}$ column.   $[\mathbf \Phi]_{i,j}$ is the importance value of the element in the $i^{th}$ row and $j^{th}$ column of  $\mathbf{W}$,  and    $\mathbf  X $ is the layer input. Following WoodFisher \cite{singh2020woodfisher} and SparseGPT \cite{frantar-sparsegpt}, the importance score reflects the minimum error  of the layer-wise outputs (in terms of $\ell_2$ norm) caused by removing a single weight.  Note that the minimum error  is evaluated by removing a single element from the weight matrix and it is not optimal in the case of simultaneously removing multiple weights. 

\textbf{Mask sharing. } Given the column and row scores, we encode the architecture information by two masks: $\mathbf{S}_{attn} \in \mathbb{R}^{M}$ for the self-attention module and $\mathbf{S}_{mlp} \in \mathbb{R}^{P}$ for the MLP module for the layers in each building block.
Different layers in the same module (self-attention or MLP) share the same mask to align the internal computations. 
We consider minimizing the collective importance scores for both the self-attention and MLP modules as below,
\begin{equation} 
    \min_{\mathbf{S}_{attn}}
    \|
    \mathbf{S}_{attn} \odot (
    \Phi_{\mathbf{W}_{Q}}^r +
    \Phi_{\mathbf{W}_{K}}^r +
    \Phi_{\mathbf{W}_{V}}^r +
    \Phi_{\mathbf{W}_{O}}^c 
    )
    \|,   
    \label{eq:init_attn_mask}
\end{equation}  
\begin{equation} 
    \min_{\mathbf{S}_{mlp}}
    \|
    {\mathbf{S}_{mlp}} \odot (
    \Phi_{\mathbf{W}_{U}}^r +
    \Phi_{\mathbf{W}_{G}}^r +
    \Phi_{\mathbf{W}_{D}}^c
    )
    \|,
    \label{eq:init_mlp_mask}
\end{equation}  
where $\| \cdot \|$ denotes the $\ell_1$ norm and $\odot$ means the element-vise multiplication.  
Given the target model size, we uniformly set the same inheriting ratio for the masks in all building blocks.  To obtain the mask in each block, we perform sorting for the sum of the corresponding scores in Equation (\ref{eq:init_attn_mask}) and (\ref{eq:init_mlp_mask}),  and inherit/keep the subnets with larger scores as the initialized architecture with the target size following the inheriting ratio, while other rows/columns with smaller scores are omitted.  

\begin{wrapfigure}{R}{0.57\textwidth}
\vspace{-1.0cm}
\begin{algorithm}[H]
        \label{alg:mask_mutation}
        \caption{Mask Mutation}
        \KwIn{
        $\mathbf{S}$, $P_m$, $\mathcal{\gamma}$, $\alpha$, $\eta$
        }
        $P_{r} \leftarrow Random(0,1)$
        
        \uIf{$Inheriting\_Ratio(\mathbf{S}) == \mathcal{\gamma} $ and $P_{r} > P_m$}{
            \KwOut{
        $\mathbf{S}$
        }
        }

        $ N  \leftarrow len( \mathbf{S} ),  iter \leftarrow 0$ 

        $Idx_{1} \leftarrow\{ \mathbf{S} == 1 \}, Idx_{2} \leftarrow\phi $
        
        \While{ $ len( Idx_{1} \cap  Idx_{2} ) < \alpha * N$ and $iter<\eta $ }{
            $ Idx_{2}   \leftarrow Random\_Subset( \{ 0,1,\cdots,N-1 \} | \gamma ) $
            
            $iter  \leftarrow iter + 1$
        }
        $\mathbf{S}^{\prime} = \mathbb{O}^N ; \ \mathbf{S}^{\prime}[Idx_2] \leftarrow 1 $
        
        \KwOut{
        $\mathbf{S}^{\prime}$ if $iter < \eta$ else $\mathbf{S}$
        }

\end{algorithm}
\vspace{-0.7cm}
\end{wrapfigure}

\subsection{Architecture Search}

In this section, we present our comprehensive training-free search framework with the visualization of the search process for one block of the LLaMA model shown in Figure~\ref{fig:search_overview}.
We first delineate the methodology for mutation based on the initialized selection masks.
Next, we define the search space and present the search pipeline.
Besides, we verify the effectiveness of our initialization strategy by comparing the convergence speeds with and without our initialization.

\subsubsection{Mask Mutation} \label{sec:mask_mutation}
During the search, we use mask mutation to generate new masks and thus new subnets to explore the search space. 
The inheriting ratio for the selection mask $\mathbf{S}_{attn} $ is denoted as $\mathbf{\Gamma}_{attn}= \{ \mathbf{\gamma}_{attn}^i \}_{i=1}^h  $ where $h$ is the number of heads, and the inheriting ratio for $\mathbf{S}_{mlp}$ is $\mathbf{\gamma}_{mlp}$. 
The mutation function  $\mathcal{M}$   with the original mask $\mathbf{S}_{attn}$ or $\mathbf{S}_{mlp}$, mutation probability $P_m$,  inheriting ratio requirement ${\gamma}_{attn}^i$ or ${\gamma}_{mlp}$, similarity ratio $\alpha$, and maximum iteration $\eta$ can be represented as follows,
\begin{equation} \footnotesize
    \mathbf{S}^{\prime}_{attn} = \{ \mathcal{M}(\mathbf{S}^i_{attn}, P_m, \mathbf{\gamma}^i_{attn}, \alpha, \eta) \}_{i=1}^h,
\end{equation}
\begin{equation} \footnotesize
    \mathbf{S}^{\prime}_{mlp} = \mathcal{M}(\mathbf{S}_{mlp}, P_m, \mathbf{\gamma}_{mlp}, \alpha, \eta),
\end{equation}
where $\mathbf{S}^i_{attn} \in \mathbb{R}^{h_m}$ denotes the selection mask for the $i^{th}$ head  and $h_m$ is the head dimension. 
In details, we show the mask mutation process with Algorithm~\ref{alg:mask_mutation}.
If the inheriting ratio of input $\mathbf{S}$  already satisfies the requirement $\gamma$ and the mutation is unnecessary based on the random generated $P_r$ (i.e.,  $P_r > P_m$), we do not mutate and simply return $\mathbf{S}$. Otherwise, given $\mathbf{S}$ and thus the set of indices $Idx_1$ for the inherited rows or columns, we try to generate a new set of indices $Idx_2$  through random sampling between $0$ to $\text{len}(\mathbf S)-1$, such that (i) $Idx_2$ follows the required inheriting ratio requirement $\gamma$, and (2) the similarity of  $Idx_1$ and  $Idx_2$ (intersection set) is larger than threshold $\alpha$. 


\begin{table}[b]
\vspace{-0.3cm}
\caption{Search space for different model sizes of OPT model family and LLaMA model family, where the notation $[a, b, c]$ specifies a range from $a$ to $b$ with an interval of $c$.}
\vspace{-0.25cm}

\centering
\resizebox{1.0 \linewidth}{!}{

\begin{tabular}{c|ccc |cccc  |cc }
\toprule
\multicolumn{1}{c|}{Model} & \multicolumn{3}{c|}{OPT Family} & \multicolumn{4}{c|}{LLaMA Family}  & \multicolumn{2}{c}{General} \\
\midrule
 Space   & \multicolumn{3}{c|}{Model Depth} & \multicolumn{4}{c|}{Model Depth}  & \multicolumn{2}{c}{Inheriting Ratio} \\ 
\midrule

  \# Params.   & 125M & 1.3B & 2.7B &  7B & 13B & 30B & 65B & $ \{\mathbf{\gamma}^i_{attn} \}_{i=1}^h $ & $\mathbf{\gamma}_{mlp}$ \\
\midrule

90\% & $[12,12,1]$ & $[24,24,1]$ & $[32,32,1]$ & $[32,32,1]$ & $[40,40,1]$ & $[60,60,1]$ & $[80,80,1]$ & $[0.9,1,0.01]$ & $[0.6,1,0.05]$ \\

80\% & $[12,12,1]$ & $[24,24,1]$ & $[30,32,1]$ & $[32,32,1]$ & $[40,40,1]$ & $[60,60,1]$ & $[80,80,1]$ & $[0.8,1,0.01]$ & $[0.4,1,0.05]$ \\

70\% & $[10,12,1]$ & $[20,24,1]$ & $[28,32,1]$ & $[30,32,1]$ & $[36,40,1]$ & $[56,60,1]$ & $[76,80,1]$ & $[0.3,1,0.01]$ & $[0.2,1,0.05]$ \\

60\% & $[10,12,1]$ & $[20,24,1]$ & $[28,32,1]$ & $[28,32,1]$ & $[32,40,1]$ & $[52,60,1]$ & $[72,80,1]$ & $[0.6,1,0.01]$ & $[0.1,1,0.05]$ \\

50\% & $[8,12,1]$ & $[16,24,1]$ & $[24,32,1]$ & $[28,32,1]$ & $[32,40,1]$ & $[52,60,1]$ & $[72,80,1]$ & $[0.6,1,0.01]$ & $[0.1,1,0.05]$ \\

\bottomrule
\end{tabular}

}

\label{tab:search_space}
\vspace{-0.3cm}
\end{table}



\subsubsection{Search Space}

We define the LLM search space with three variables  for each transformer building block below: the model depth $d$, inheriting ratios $\mathbf{\Gamma}_{attn} = \{ \mathbf{\gamma}_{attn}^i \}_{i=1}^h$  for $\mathbf{S}_{attn}$, and $\mathbf{\gamma}_{mlp}$ for $\mathbf{S}_{mlp}$. The specifications of this search space, including the range for each factor, are detailed in Table~\ref{tab:search_space}.

$\mathbf{\gamma}_{mlp}$   has a larger search space than  $ \{ \mathbf{\gamma}_{attn}^i \}_{i=1}^h$ according to our ablation study illustrated in  Figure~\ref{fig:ablation_sparsity_attn_mlp}. 
Results are evaluated using LLaMA-7B on the WikiText2 dataset with a sequence length of 2048.
Specifically, we apply the same  local inheriting ratio for three cases, (i) the attention module only, (ii) the MLP module only, and (iii) both modules. Note that in case (i) or (ii), the global inheriting ratio is larger than case (iii) since the MLP in case (i) or the attention in case (ii) directly uses the original layers with 100\% inheriting ratio.    
From Figure~\ref{fig:ablation_sparsity_attn_mlp}, we observe that case (ii) achieves a better perplexity with a lower global inheriting ratio than case (i), demonstrating that the MLP   exhibits greater redundancy and is less sensitive to parameter reduction than the self-attention module. 
Therefore, we set a larger search space of inheriting ratios for  MLP  than  the self-attention module.

Different from other transformer-based search works~\cite{autoFormer, gong2022nasvit,liao2021searching}, we do not search the number of heads in  self-attention.
It stems from the nature of transformers that all heads are essential for representing the input data in the attention mechanism.
Moreover, we refrain from conducting searches on the embedding and output layers of LLMs, as their weights constitute only a minor fraction of the total parameters yet are vital for the precise representation of tokens.

\subsubsection{Search Pipeline} \label{sec:pipeline}

We implement our evolutionary search across the OPT and LLaMA model families with varying model sizes to derive   efficient LLM architectures/subnets. The pipeline is shown below. 


\textbf{Initial  generation.}
Given the single initialized subnet (Section \ref{sec:intialization}),  multiple candidates ($N$ subnets in total) are generated by mutation of the inheriting ratios with probability $P_s^0$  and then mask mutation (Section \ref{sec:mask_mutation}) with probability $P_m^0$. The depth mutation  is not involved at this initial step. The top $k$ subnets are preserved as the initial  generation. 

\textbf{Following  generation.}
With the $k$ subnets as parental candidates, a new population with $N$ candidates are generated through   mutation and crossover.  We select a random parental candidate for mutation until  the number of mutation candidates reaches a threshold $N_m$. 
Mutation involves altering the depth with  probability  $P_d$, altering the  inheriting ratios with  probability  $P_s < P_s^0$, and mutating the mask with  probability  $P_m < P_m^0$ (see Algorithm \ref{alg:mask_mutation}).  The probabilities are smaller than initial generation as superior candidates should be preserved with less randomness. 
For the crossover, two  parental candidates are randomly selected and combined to form a new candidate until there are $N_c$ candidates. With the population from the parental candidates, top $k$ subnets are preserved as the next generation. 

\textbf{Candidate evaluation.} 
For each generated candidate, if its parameter number does not fall in the range of the target model size,  it is  unsatisfying and we simply drop it.  To compare  candidate subnets, we evaluate them with a few random training samples  from WikiText2 to compute the perplexity.

\begin{figure*}[t]
\vspace{-0.5cm}
\begin{minipage}{0.505\textwidth}

  \centering
  \includegraphics[width=1.0\textwidth]{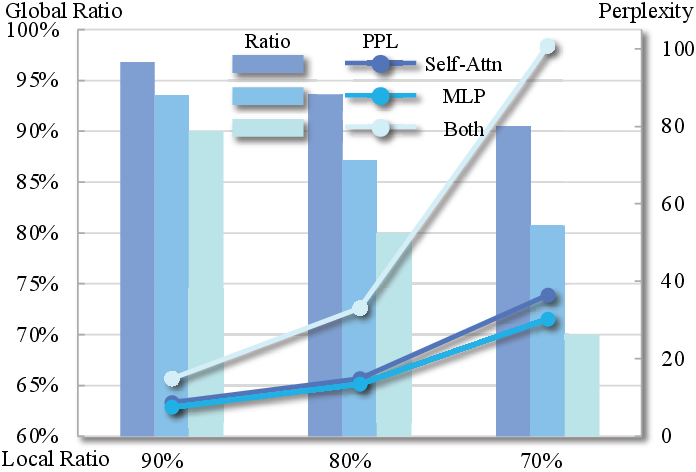}
  \vspace{-0.7cm}
  \caption{
  Ablation analysis of the inheriting ratios applied to the self-attention, MLP, or both.
  }
  \label{fig:ablation_sparsity_attn_mlp}
  \end{minipage}\hfill
\begin{minipage}{0.465\textwidth}
\centering
  \includegraphics[width=1.0\textwidth]{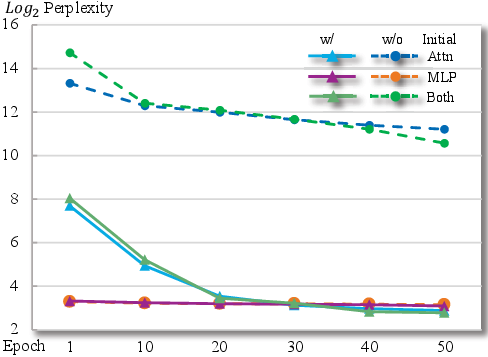}
  \vspace{-0.7cm}
  \caption{
  Ablation analysis of convergence speed with or without our initialization.
  }
  \label{fig:ablation_search_process}
\end{minipage}
\vspace{-0.4cm}
\end{figure*}

\textbf{Necessity for initialization.} 
To verify the effectiveness of our initialization in the search,  we ablate the initialization for three cases, (i) self-attention only, (ii) MLP  only, and (iii) both modules. LLaMA-7B is adopted with an 80\% inheriting ratio for selection masks. Results are evaluated on   Wikitext2  with a 2048 sequence length. As shown in Figure~\ref{fig:ablation_search_process},  the self-attention module complicates the identification of an effective subnet without our initialization strategy.
In contrast, the MLP module exhibits less sensitivity to initialization.  The search in both modules struggles to yield effective subnets without our initialization, primarily due to self-attention. The observation underscores the necessity of our initialization approach.




\subsection{Reformation}

After the search, we can obtain a subnet from the original LLM. To  improve the subnet  performance, we further reform the weights in the subnet by using the omitted weights  to compensate their loss. Specifically,  for each linear layer in the subnet with their   original  weights $\mathbf W$ before the search,   we would like to reform the weights under    the searched mask $\mathbf M$ and obtain    $\widehat{\mathbf W}$,  so that the layer output difference in  $\ell_2$ norm, i.e., 
 $\| \widehat{ \mathbf W} \mathbf X - \mathbf W \mathbf X  \|_2^2  $, is minimized.
The problem is formulated below, 
{\small
\begin{align} \label{eq:ori_problem}
\min_{\widehat{ \mathbf W}} \ \ &   \| \widehat{ \mathbf W} \mathbf X - \mathbf W \mathbf X  \|_2^2 , \nonumber  \\ 
\text{s.t.} \ \ 
&    \widehat{ \mathbf W} \odot  \mathbf{M}   =  \mathbf 0, 
\end{align}}%
where  $\mathbf{M}$ indicates the location of pruned weights with element 1 denoting pruned and 0 denoting unpruned weights. 
Here we only reform inherited columns based on omitted columns in $\mathbf{W}$ rather than reforming rows with omitted rows, since the output corresponding to omitted rows are always zeros which are unavailable for any compensations by modifications in other rows. 
To solve this problem, we propose a solution based on alternating direction method of multipliers (ADMM) \cite{boyd2011distributed,parikh2014proximal,goldstein2014fast}  with the following theorem.  The  detailed proof is shown  in Appendix~\ref{app:sec:proof}. 
\begin{theorem}
Problem~(\ref{eq:ori_problem}) can be solved in iterations. In the $k^{th}$ iteration, it  performs the   updates:
{\small
\begin{align}  \label{eq:optimal_w}
\widehat{ \mathbf W}^{k+1} = & (\mathbf{X} \mathbf{X}^T + \rho \mathbf{I})^{-1} (\mathbf{X} \mathbf{X}^T \mathbf{ W}^T + \rho (\mathbf Z^k  - \mathbf U^k )^T),  \\
\mathbf Z^{k+1} = & \left( \widehat{ \mathbf W}^{k+1} +  \mathbf U^k \right) \odot \mathbf{M},  \\
\mathbf U^{k+1} =  & \mathbf U^k + \mathbf W^{k+1} - \mathbf Z^{k+1}, 
\end{align}}%
where $\rho > 0$  is  the penalty parameter. The initial variable values at $k=0$ follow the configurations that  $\widehat{ \mathbf W}^0 = \mathbf W$, $\mathbf Z^0 = \widehat{ \mathbf W}^0$, and $\mathbf U^{0} = \mathbf 0$.
\end{theorem}

In practice, we find that after a few tens iterations (such as 20 or 30), the loss in  Problem~(\ref{eq:ori_problem}) converges.  The  complexity is determined by the inversion operation, which is the same as SparseGPT~\cite{frantar-sparsegpt}.  However, as it can converge  fast within 30 iterations,  our ADMM-based solution typically requires less computation time than SparseGPT which needs to iterate over  all   rows in the weight matrix.

\subsection{Efficient Inference} \label{sec:small_dense}

After searching and reformation, we can get optimal efficient subnets with the selection masks $\mathbf{S}_{attn} \in \mathbb{R}^{M}$ and $\mathbf{S}_{mlp} \in 
\mathbb{R}^P $ for each block of the LLMs. 
We further convert the subnets into small-dense models following the masks for efficient inference. Thus, the dimension of the weight is actually reduced with faster inference speed.  More details can be found in Appendix \ref{app:sec:small_dense}.

\begin{table}[t]
\vspace{-0.5cm}
\centering
\caption{
Results of the compressed LLaMA-7B and LLaMA-13B on the WikiText2 dataset, PTB dataset, and other common sense reasoning datasets.
The perplexity on the WikiText2 and PTB is calculated with the 2048 sequence length.
The accuracy results are evaluated with the same pipeline as LLM-Pruner~\cite{ma2023llmpruner} to ensure a fair comparison.
The average is computed across seven classification datasets.
LLM-Pruner (v), (e2). and (e1) denote the vector-wise and element-wise importance, (c) and (b) denote the channel and block strategies.
}
\vspace{-0.25cm}
\label{tab:main_results}
\resizebox{1.0\linewidth}{!}{
\begin{tabular}{c|c|cc|ccccccc|c}

\toprule
\multirow{2}{*}{Method} & Inheriting & Wiki    & PTB     & \multirow{2}{*}{BoolQ} & \multirow{2}{*}{PIQA}  & Hella & Wino & \multirow{2}{*}{ARC-e}          & \multirow{2}{*}{ARC-c}  & \multirow{2}{*}{OBQA}  & Average         \\
 & Ratio & PPL$\downarrow$ & PPL$\downarrow$ & & & Swag & Grande & & & & Acc.$\uparrow$ \\
\toprule
LLaMA-7B  & 100\%      & 5.68          & 27.34          & 73.18 & 78.35 & 72.99           & 67.01      & 67.45          & 41.38       & 42.40 & 63.25           \\
\toprule
LLM-Pruner(v)     & \multirow{3}{*}{90\%}     & 7.73         & 38.94         & 67.95 & 77.42 & 69.31           & 63.54      & 66.33          & 39.85       & 41.20 & 60.80          \\
LLM-Pruner(e2)   &      & 7.46         & 36.87          & 68.29 & 76.88 & 70.25           & 64.33      & 65.28          & 40.10        & 39.60 & 60.68          \\
LLM-Pruner(e1)   &      & 7.42         & 36.73         & 66.97 & 77.26 & 70.30            & 64.33      & 65.24          & 40.19       & 41.00   & 60.76          \\
                            \midrule
SliceGPT         &   90\%   & 7.00        & 133.80       & 57.68 & 69.80  & 59.32           & 68.11      & 62.75          & 36.01       & 38.00   & 55.95           \\
                            \midrule
FLAP             &   90\%   & 6.34          & 32.39          & 74.43 & 75.41 & 68.68           & 67.01      & 65.78          & 38.48       & 41.00   & 61.54          \\
\midrule
Ours          &   90\%   & \textbf{6.10}  & \textbf{32.05} & 74.37 & 76.88 & 70.71           & 67.56      & 68.39          & 40.10        & 39.20 & \textbf{62.46} \\
\toprule
LLM-Pruner(v)         & \multirow{3}{*}{80\%}     & 10.73         & 59.73         & 61.44 & 71.71 & 57.27           & 54.22      & 55.77          & 33.96       & 38.40 & 53.25           \\
LLM-Pruner(e2)        &      & 11.97         & 55.68         & 59.39 & 75.57 & 65.34           & 61.33      & 59.18          & 37.12       & 39.80 & 56.82           \\
LLM-Pruner(e1)       &      & 10.73         & 59.73         & 57.06 & 75.68 & 66.80            & 59.83      & 60.94          & 36.52       & 40.00   & 56.69           \\
                            \midrule
SliceGPT      & 80\%     & 8.71        & 143.89        & 37.89 & 64.09 & 45.67           & 62.75      & 53.62          & 31.74       & 33.20 & 46.99          \\
                            \midrule
FLAP           & 80\%     & 7.40           & 36.77          & 68.59 & 74.21 & 64.98           & 64.40       & 59.89          & 37.80        & 40.20 & 58.58           \\
\midrule
Ours            & 80\%     & \textbf{6.89} & \textbf{36.06} & 70.98 & 74.92 & 67.29           & 64.64      & 64.23          & 36.52       & 39.40 & \textbf{59.71}  \\
\bottomrule
\toprule
LLaMA-13B                & 100\%      & 5.09          & 19.23          & 68.47 & 78.89 & 76.24           & 70.09      & 74.58          & 44.54       & 42.00   & 64.97           \\
\toprule
LLM-Pruner(c)     & \multirow{2}{*}{90\%}     & 7.70           & 35.32          & 68.47 & 74.76 & 66.99           & 66.38      & 66.58          & 35.24       & 38.20 & 59.52           \\
LLM-Pruner(b)    &      & 6.38          & 31.85          & 70.64 & 78.40  & 75.00           & 69.46      & \textit{72.82} & 41.47       & 41.40 & 64.17           \\
                            \midrule
SliceGPT     & 90\%     & 6.43          & 86.09          & 61.74 & 69.97 & 60.74           & 69.38      & 66.79          & 40.70        & 41.80 & 58.73           \\
                            \midrule
FLAP            & 90\%     & 5.45        & 20.98        & 63.76 & 78.07 & 73.69           & 69.61      & 69.53          & 39.93       & 41.60 & 62.31           \\
\midrule
Ours            & 90\%     & \textbf{5.39} & \textbf{20.63} & 71.65 & 78.18 & 75.04           & 69.61      & 69.70           & 43.09       & 42.60 & \textbf{64.23}  \\
\toprule
LLM-Pruner(c)    & \multirow{2}{*}{80\%}     & 48.76         & 218.49         & 62.39 & 66.87 & 49.17           & 58.96      & 49.62          & 31.83       & 33.20 & 50.29           \\
LLM-Pruner(b)    &      & 10.05         & 55.46          & 67.68 & 77.15 & 73.41           & 65.11      & 68.35          & 38.40        & 42.40 & 61.79           \\
                            \midrule
SliceGPT         & 80\%     & 7.55          & 117.94         & 50.34 & 66.00    & 53.37           & 68.11      & 60.56          & 36.35       & 38.20 & 53.27           \\
                            \midrule
FLAP       & 80\%     & 6.03        & 23.33         & 62.23 & 76.50  & 70.59           & 68.35      & 65.66          & 38.99       & 41.60 & 60.56           \\
\midrule
Ours       & 80\%     & \textbf{5.90}  & \textbf{22.66} & 68.53 & 77.09 & 72.60            & 69.22      & 66.25          & 40.02       & 41.00   & \textbf{62.10} \\

\bottomrule

\end{tabular}
}
\vspace{-0.6cm}
\end{table}

\section{Experiments}

\subsection{Experiment Settings}

\textbf{Hyper-parameter setting.}  For the evolutionary search, we adopt specific hyper-parameters as follows: the population size ($N$), the number of mutations ($N_m$), and the number of crossovers ($N_c$) are  set to 100, 50, and 30, respectively.
In each generation, the top $10$ subnets are selected as parental candidates to produce offspring networks through the mechanisms of mutation and crossover.
The rest subnets in the population are generated with mutation with larger randomness (i.e., same as initial mutation).
The initial mutation probabilities ($P_m^0$ and $P_s^0$) are set at 0.6 and 0.3 to promote variability early in the search process.
Subsequently, for ongoing generation processes, the mutation probabilities ($P_m$ and $P_s$) are adjusted to 0.3 and 0.1, while the probability for depth ($P_d$) is maintained at 0.1.
The similarity ratio $\alpha$ and maximum iteration $\eta$ are set at 0.8 and 1000 in mask mutation.
The total evolution epoch is 50.
For the reformation, we adopt $\rho$ as 1.0 and the iteration number as 30.


\textbf{Datasets and metrics.} We compare the perplexity of the models on the WikiText2~\cite{wikitextdataset} and PTB~\cite{ptbdataset} datasets with the 2048 sequence length. We also compare the zero-shot accuracy on  common reasoning zero-shot classification datasets including BoolQ~\cite{clark2019boolq}, PIQA~\cite{bisk2020piqa}, HellaSwag~\cite{zellers2019hellaswag}, WinoGrande~\cite{sakaguchi2021winogrande}, ARC-easy~\cite{clark2018arc}, ARC-challenge~\cite{clark2018arc}, and OpenbookQA~\cite{mihaylov-etal-2018-suit-opqa}.

\textbf{Models.} We evaluate on multiple LLM families including  LLaMA \cite{touvron2023llama}, Vicuna \cite{zheng2023vicuna} and OPT \cite{zhang2022opt}. 

\textbf{Baselines and pipeline. }  
We compare with SOTA baselines including LLM-pruner \cite{ma2023llmpruner}, SliceGPT \cite{ashkboos2024slicegpt} and FLAP \cite{an2023flap}. 
We adhere to the exact evaluation pipeline from the well-known LLM-Pruner~\cite{ma2023llmpruner}, which is applied  for all approaches to ensure a fair comparison. 
For the reformation, we randomly select 128 samples from training split of WikiText2, with the same random seed and thus same data for the calibration of other works, including SliceGPT and FLAP, ensuring a fair comparison.

\textbf{Search Cost. }
We leverage the evolution search on NVIDIA A100 40G GPUs.
Specifically, to explore the subnets of LLaMA-7B, we finish the search on one GPU with around 5 hours.




\begin{table}[t]
\vspace{-0.7cm}
\centering
\caption{
Results of compressed Vicuna-7B.
}
\vspace{-0.25cm}
\label{tab:results_vicuna_7b}
\resizebox{1.0\linewidth}{!}{
\begin{tabular}{c|c|cc|ccccccc|c}
\toprule
\multirow{2}{*}{Method} & Inheriting & Wiki          & PTB            & \multirow{2}{*}{BoolQ} & \multirow{2}{*}{PIQA} & Hella & Wino   & \multirow{2}{*}{ARC-e} & \multirow{2}{*}{ARC-c} & \multirow{2}{*}{OBQA} & Average        \\
                        &            Ratio               & PPL$\downarrow$           & PPL$\downarrow$            &                        &                       & Swag  & Grande &                        &                        &                       & Acc.$\uparrow$           \\
                        \toprule
Vicuna-7B               & 100\%          & 6.78          & 26.78          & 76.57                  & 77.75                 & 70.64 & 67.40  & 65.11                  & 41.21                  & 40.80                 & 62.78          \\
\toprule
LLM-Pruner(e2)          &   \multirow{4}{*}{90\%}                   & 8.74          & 30.69          & 61.68                  & 75.95                 & 69.35 & 64.72  & 68.86                  & 39.16                  & 40.00                 & 59.96          \\
SliceGPT                &                      & 8.23          & 61.83          & 67.00                  & 69.64                 & 58.65 & 63.77  & 54.59                  & 37.71                  & 38.40                 & 55.68          \\
FLAP                    &                     & 7.64          & 29.95          & 74.43                  & 75.41                 & 68.68 & 67.01  & 65.78                  & 38.48                  & 41.00                 & 61.54          \\
Ours                    &                      & \textbf{7.18} & \textbf{28.87} & 75.10                  & 75.90                 & 69.97 & 66.92  & 67.11                  & 40.61                  & 40.40                 & \textbf{62.29} \\
\midrule
LLM-Pruner(e2)          & \multirow{4}{*}{80\%}                     & 12.97         & 46.34          & 62.87                  & 75.41                 & 64.00 & 58.41  & 60.98                  & 37.12                  & 39.00                 & 56.83          \\
SliceGPT                &                      & 10.13         & 94.93          & 53.82                  & 64.42                 & 49.14 & 60.38  & 52.27                  & 34.56                  & 33.40                 & 49.71          \\
FLAP                    &                      & 8.90          & 34.04          & 69.14                  & 72.52                 & 63.01 & 64.48  & 61.11                  & 34.56                  & 39.00                 & 57.69          \\
Ours                    &                      & \textbf{8.20} & \textbf{33.59} & 66.30                  & 74.92                 & 65.05 & 62.90  & 64.63                  & 38.91                  & 39.80                 & \textbf{58.93} \\
\bottomrule
\end{tabular}
}
\vspace{-0.3cm}
\end{table}



\begin{table}[t]
\small

\begin{minipage}{0.45\linewidth}
\caption{
Results of compressed OPT.
}
\vspace{-0.25cm}
\label{tab:results_opt_models}
\centering
\resizebox{1.0\linewidth}{!}{
\setlength{\tabcolsep}{1.5pt} 
\begin{tabular}{c|cc|cc|cc}
\toprule
Ratio & \multicolumn{2}{c|}{90\%}        & \multicolumn{2}{c|}{80\%}        & \multicolumn{2}{c}{70\%}         \\
\midrule
 Method         & Wiki  $\downarrow$         & PTB        $\downarrow$    & Wiki  $\downarrow$         & PTB  $\downarrow$          & Wiki    $\downarrow$       & PTB  $\downarrow$           \\
\midrule
\multicolumn{7}{c}{OPT-125M Wiki: 27.65 PTB: 38.99} \\
\midrule
 SliceGPT       & 35.31          & 75.59          & 54.88          & 149.17         & 84.16          & 245.18          \\
         Ours           & \textbf{30.97} & \textbf{40.14} & \textbf{44.12} & \textbf{66.55} & \textbf{80.84} & \textbf{124.27} \\
\midrule
\multicolumn{7}{c}{OPT-1.3B Wiki: 14.63 PTB: 20.29 } \\
\midrule
SliceGPT       & 16.74          & 35.31          & 20.17          & 61.30          & 28.53          & 113.42          \\
        Ours           & \textbf{15.51} & \textbf{20.19} & \textbf{19.23} & \textbf{32.81} & \textbf{26.82} & \textbf{69.42}  \\
\midrule
\multicolumn{7}{c}{OPT-2.7B Wiki: 12.47 PTB: 17.97 } \\
\midrule
SliceGPT       & 14.10          & 37.01          & 16.81          & 65.09          & 24.12          & 132.13          \\
        Ours           & \textbf{13.32} & \textbf{17.24} & \textbf{16.44} & \textbf{23.66} & \textbf{23.48} & \textbf{58.46} \\
\bottomrule
\end{tabular}
}
\end{minipage}
	\begin{minipage}{0.55\linewidth}
        \centering
\caption{
Results with lower inheriting ratio.
}
\vspace{-0.25cm}
\label{tab:results_larger_sparsity_llama_7b_13b}
\centering
\resizebox{1.0\linewidth}{!}{
\setlength{\tabcolsep}{1.5pt} 
\begin{tabular}{c|cc|cc|cc}
\toprule
Ratio         & \multicolumn{2}{c|}{70\%}       & \multicolumn{2}{c|}{60\%}        & \multicolumn{2}{c}{50\%}         \\
\midrule
Method         & Wiki $\downarrow$         & PTB $\downarrow$           & Wiki  $\downarrow$         & PTB     $\downarrow$       & Wiki $\downarrow$           & PTB $\downarrow$            \\
\midrule
\multicolumn{7}{c}{LLaMA-7B} \\
\midrule
LLM-Pruner(e2) & 18.58         & 93.24          & 38.27          & 238.09         & 125.96         & 460.73          \\
                     SliceGPT       & 15.95         & 583.58         & 279.52         & 5186.06        & 1830.43        & 15333.66        \\
                     FLAP           & 9.18          & 47.35          & 12.34          & 65.54          & 21.89          & 135.84          \\
                     Ours           & \textbf{8.28} & \textbf{45.26} & \textbf{10.21} & \textbf{62.07} & \textbf{15.48} & \textbf{117.06} \\
\midrule                  
\multicolumn{7}{c}{LLaMA-13B} \\
\midrule
LLM-Pruner(e2) & 22.36         & 112.03         & 66.38          & 278.36         & 3827.63        & 1287.11         \\
                     SliceGPT       & 9.79          & 167.27         & 13.21          & 247.71        & 19.95          & 408.68          \\
                     FLAP           & 6.97          & 27.38          & 8.67           & 35.91         & 12.88          & 53.54           \\
                     Ours           & \textbf{6.67} & \textbf{26.37} & \textbf{8.00}  & \textbf{33.23} & \textbf{10.44} & \textbf{45.73} \\
                     \bottomrule
\end{tabular}
}
\end{minipage}\hfill

\vspace{-0.5cm}
\end{table}



\subsection{Main Results}


\textbf{Superior performance compared with SOTA baselines.  } We show our  results of  LLaMA-7B and LLaMA-13B in Table~\ref{tab:main_results} and Figure~\ref{fig:sota-visual} (b) and (c).
We observe that our method outperforms all baselines in terms of perplexity on WikiText2 and PTB, and average zero-shot accuracy (over multiple zero-shot datasets). 
Take LLaMA-13B on WikiText2 as an example, our method improves the perplexity by 4.15, 1.65, and 0.13 compared to LLM-Pruner(b), SliceGPT, and FLAP, respectively, with a 80\% inheriting ratio. 
Meanwhile, it achieves a higher average accuracy on   seven classification datasets than   baselines.  For instance, under a 80\% inheriting ratio, our method on LLaMA-7B improves the average accuracy by 2.89\%, 12.72\%, and 1.13\% compared to LLM-Pruner(e1), SliceGPT, and FLAP, respectively.
As the SliceGPT is sensitive to the calibration dataset, we further show the results with PTB calibration in Appendix~\ref{app:sec:slicegpt_ptb_calibrate}. Besides, we ablate the search with PTB in Appendix~\ref{app:sec:search_on_ptb}.

The comparisons on other LLM families are demonstrated in Table \ref{tab:results_vicuna_7b} for  Vicuna-7B~\cite{zheng2023vicuna} and Table~\ref{tab:results_opt_models} with Figure~\ref{fig:sota-visual} (a) for OPT family~\cite{zhang2022opt}.  As LLM-Pruner and FLAP do not implement their methods on  OPT, we only compare with SliceGPT for OPT models.  We can make similar observations that our method performs the best compared with SOTA baselines, demonstrating a superior generalization performance across various  datasets/tasks, model  structures, and inheriting ratios. 

\textbf{Scaling to  small inheriting ratios and large models. }  Furthermore, our method consistently performs the best when scaling to larger model sizes or smaller inheriting ratios, indicating the great potential of our method  for the ever-increasing LLM model size.  Specifically, we   show the results of LLaMA-7B and LLaMA-13B with lower inheriting ratios and thus more efficient   models in  Table~\ref{tab:results_larger_sparsity_llama_7b_13b} and Figure~\ref{fig:sota-visual} (b) and (c).  Our method consistently performs the best with more significant improvements under smaller inheriting ratios such as 50\%. 
Besides, we deliver results on large models including LLaMA-30B and LLaMA-65B in Table~\ref{tab:results_extra_large_llama_30b_65b} and Figure~\ref{fig:sota-visual} (d). Our method achieves superior performance than FLAP under various settings, verifying our effectiveness and generalization.

\begin{table}[]
\vspace{-0.7cm}
\centering
\caption{
Results of extra large LLaMA models.
}
\vspace{-0.25cm}
\label{tab:results_extra_large_llama_30b_65b}
\resizebox{1\linewidth}{!}{

\begin{tabular}{c|c|c|ccccc|c|ccccc}
\toprule
\multicolumn{2}{c|}{Model}        & \multicolumn{6}{c|}{LLaMA-30B}               & \multicolumn{6}{c}{LLaMA-65B}                 \\
\midrule
Dataset               & Ratio & 100\%   & 90\%  & 80\% & 70\% & 60\%  & 50\%  & 100\%   & 90\%  & 80\%  & 70\%  & 60\%  & 50\%  \\
\midrule
\multirow{2}{*}{Wiki$\downarrow$} & FLAP     & \multirow{2}{*}{4.10}   &   4.52    &   5.18   &   6.28   &    8.73   &   13.41    &  \multirow{2}{*}{3.53}     &   3.91    &   4.45    &   5.10    &    6.16   &   8.11    \\
                      & Ours     &    & \textbf{4.44}  & \textbf{4.94} & \textbf{5.63} & \textbf{6.70}   & \textbf{8.01}  &   & \textbf{3.84}  & \textbf{4.29}  & \textbf{4.80}   & \textbf{5.49}  & \textbf{6.45}  \\
                      \midrule
\multirow{2}{*}{PTB$\downarrow$}  & FLAP     & \multirow{2}{*}{16.29} &    17.29   &  19.30    &   21.88   &    29.11   &    47.30   &       \multirow{2}{*}{17.61} &    19.35   &   21.01    &   22.45    &   25.97    &   33.86    \\
                      & Ours     &  & \textbf{17.15} & \textbf{18.80} & \textbf{20.70} & \textbf{24.22} & \textbf{30.51} &  & \textbf{19.22} & \textbf{20.40} & \textbf{21.39} & \textbf{23.41} & \textbf{30.08} \\
                      \bottomrule
\end{tabular}
}
\vspace{-0.4cm}
\end{table}

\subsection{Ablation Study} \label{sec:ablation}




\begin{wraptable}{r}{0.5\textwidth}
\vspace{-0.7cm}
\centering
\caption{
LLaMA-7B perplexity ($\downarrow$) results on WikiText2 dataset with 128 sequence length.
}
\label{tab:llama_7b_128_length_results}
\resizebox{1.0\linewidth}{!}{
\begin{tabular}{c|cccc}
\toprule
Ratio & LLM-Pruner(e1) & SliceGPT & FLAP                      & Ours  \\
\midrule
90\%     & 15.22          & 14.18    & 14.15 & \textbf{13.40} \\
80\%     & 19.09          & 17.08    & 14.62                     & \textbf{14.54} \\
70\%     & 30.63          & 24.39    & 17.62                     & \textbf{17.11} \\
60\%     & 52.30          & 40.04    & 23.53 & \textbf{20.24} \\
50\%     & 106.07         & 74.09    & 31.80                     & \textbf{26.96} \\
\bottomrule
\end{tabular}
}
\vspace{-0.5cm}
\end{wraptable}


We show the results of LLaMA-7B with 128 sequence length in Table~\ref{tab:llama_7b_128_length_results}.
Our method performs the best across different inheriting ratios, indicating the effectiveness on short sequences.
Results of LLaMA-13B are in Appendix~\ref{app:sec:ablation_128_llama13b}.

\begin{figure*}[h]
\begin{minipage}{0.485\textwidth}
  \centering
  \includegraphics[width=1.0\textwidth]{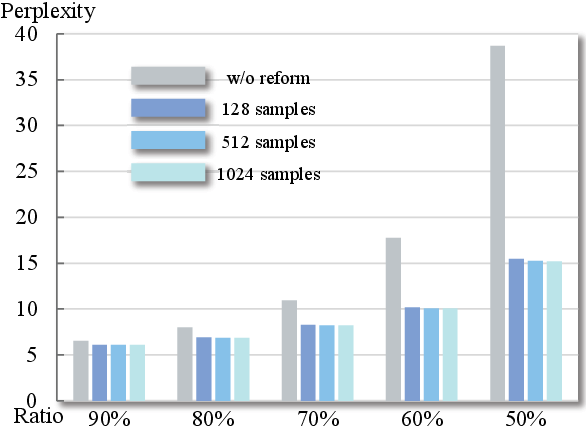}
  \vspace{-0.7cm}
  \caption{
  Ablation analysis for reformation with different numbers of samples.
  }
  \label{fig:ablation_reformation}
  \end{minipage}\hfill
\begin{minipage}{0.485\textwidth}
\centering
  \includegraphics[width=1.0\textwidth]{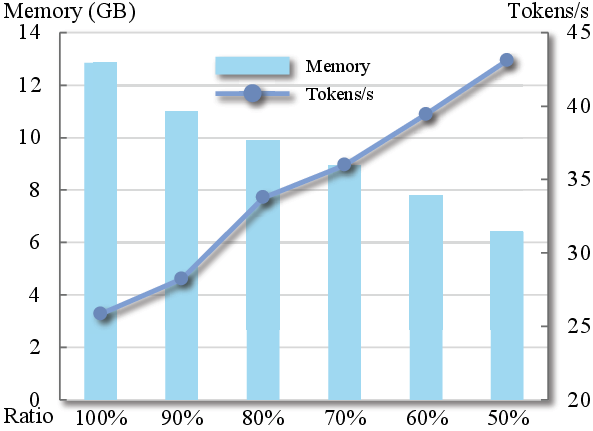}
  \vspace{-0.7cm}
  \caption{
  Analysis for memory and generation speed of LLaMA-7B on NVIDIA A100 40G.
  }
  \label{fig:gpu_memory_inference_speed}
\end{minipage}
\vspace{-0.4cm}
\end{figure*}

Besides, to verify the influence of the number of examples for the reformation, we conduct  ablation studies by varying the number from 128 to 512 and 1024. As shown in Figure~\ref{fig:ablation_reformation},  
our reformation effectively improves the performance, and it is not sensitive to the number of samples. 128 samples already provide satisfying performance.  
Furthermore, we investigate the impact of the step number  and $\rho$ in the ADMM solution for  the reformation, with detailed ablation results presented in Appendix~\ref{app:sec:ablation_reformation}.

\subsection{Generation Acceleration}

To demonstrate our acceleration performance, we report the memory consumption and inference speed with our searched LLaMA-7B models on NVIDIA A100 40G GPUs across different inheriting ratios. As shown in Figure~\ref{fig:gpu_memory_inference_speed},  we can observe that with a smaller inheriting ratio, our searched efficient model consumes less memory with a faster generation speed.

\section{Conclusion and Limitation}

In this paper, we propose a training-free search framework to find the optimal subnets inside   LLMs.
We further propose a reformation algorithm that reconstructs the weights of subnets to enhance the task performance.
The experiments show the effectiveness of our proposed method compared to SOTA structured pruning methods.
Additionally, we achieve memory reduction and practical inference acceleration on GPUs, which shows the efficiency of our method.
The search cost required by our method can increase with the model size, which takes more time for large models.

\section*{Acknowledgment}

We would like to express our sincere gratitude to Professor Lin and Professor Wang for their invaluable guidance throughout the development of this paper. 
We are also deeply grateful to Pu, Yifan, and Zhenglun for their dedicated contributions, thoughtful insights, and collaborative efforts, which were essential to the completion of this research.

\bibliographystyle{unsrt}
\bibliography{reference}

\newpage
\appendix

\onecolumn

\setcounter{table}{0}
\renewcommand{\thetable}{A\arabic{table}}
\renewcommand*{\theHtable}{\thetable}

\setcounter{figure}{0}
\renewcommand{\thefigure}{A\arabic{figure}}
\renewcommand*{\theHfigure}{\thefigure}

\noindent\textbf{\large Appendix}

\section{Efficient Inference} \label{app:sec:small_dense}

After searching and reformation, we can get optimal efficient subnets with the selections masks $\mathbf{S}_{attn} \in \mathbb{R}^{M}$ and $\mathbf{S}_{mlp} \in 
\mathbb{R}^P $ for each block of the LLMs.
In detail, for the weights of query, key, and value denoted as $\mathbf{W}_Q, \mathbf{W}_K, \mathbf{W}_V \in \mathbb{R}^{M \times D}$, we generate the weight subsets by extracting the selected rows from the original weights, which are denoted as $\mathbf{W}_Q^{\prime}, \mathbf{W}_K^{\prime}, \mathbf{W}_V^{\prime} \in \mathbb{R}^{m \times D} $.
For the weights of the output projection $\mathbf{W}_O \in \mathbb{R}^{D \times M} $, we extract the columns instead of rows and reform the selected weight based on the omitted ones for the weight subnets $\mathbf{W}_O^{\prime} \in \mathbb{R}^{D \times m} $.
Subsequently, the given input $\mathbf{X} \in \mathbb{R}^{BN \times D} $ is projected in the self-attention module as follows,
\begin{equation} \footnotesize
    \mathbf{X}^{\prime} = 
    \mathbf{W}_{O}^{\prime}  
    {\{
    softmax{[
    {(\mathbf{W}_{Q}^{\prime}   \mathbf{X})} \cdot {(\mathbf{W}_{K}^{\prime}\mathbf{X})}^T
    ]}
    \cdot
    {(\mathbf{W}_{V}^{\prime}   \mathbf{X})}
    \}}
    \in \mathbb{R}^{BN \times D}
\end{equation}

For the MLP module in the LLaMA family, we denote the weights of the three linear layers with $\mathbf{W}_{U}, \mathbf{W}_{G} \in \mathbb{R}^{P \times D} $, and $\mathbf{W}_{D} \in \mathbb{R}^{D \times P}$ for the up, gate, and down projections, respectively.
The weight subsets generated with the selection mask $\mathbf{S}_{mlp}$ for three linear layers are $\mathbf{W}_{U}^{\prime}, \mathbf{W}_{G}^{\prime} \in \mathbb{R}^{p \times D} $, and $\mathbf{W}_{D}^{\prime} \in \mathbb{R}^{D \times p}$, where only $\mathbf{W}_{D}$ is reformed.
Then, the given input $\mathbf{X} \in \mathbb{R}^{BN \times D} $ is projected in the MLP module as follows,
\begin{equation} \footnotesize
    \mathbf{X}^{\prime} = 
    \mathbf{W}_{D}^{\prime}  
    \{
    (\mathbf{W}_{U}^{\prime}   \mathbf{X})
    \odot
    activation[
    (\mathbf{W}_{G}^{\prime}   \mathbf{X})
    ]
    \}
    \in \mathbb{R}^{BN \times D}
\end{equation}
where the activation function for the LLaMA family is SiLU~\cite{silu_act}.

Therefore, the computation cost is reduced for both self-attention and MLP modules, while the configuration of the input $ \mathbf{X} \in \mathbb{R}^{BN \times D}$ is preserved for preventing information loss and maintaining the representation capability of tokens.

\section{Proof of Theorem 3.1}
\label{app:sec:proof}

Problem (\ref{eq:ori_problem}) can be reformulated as follows,
\begin{align}  \label{eq:admm_problem}
\min_{\widehat{ \mathbf W}, \mathbf Z} \ \ &   \| \widehat{ \mathbf W} \mathbf X - \mathbf W \mathbf X  \|_2^2 + g(\mathbf Z), \nonumber  \\ 
\text{s.t.} \ \ 
&    \widehat{ \mathbf W} = \mathbf Z, 
\end{align}
where $ g(\mathbf Z)$ is a inidicator function as the following, 
\begin{align}  
g(\mathbf Z) =\begin{cases}
  \begin{array}{ c l }
    \infty, & \quad  \textrm{otherwise}, \\
    0,                 & \quad  \textrm{if} \quad \widehat{ \mathbf W} \odot  \mathbf{M}   =  \mathbf 0. 
  \end{array} \end{cases}
\end{align}
We can see that Problem (\ref{eq:admm_problem}) is equvilant to Problem (\ref{eq:ori_problem}).  

Based on ADMM \cite{parikh2014proximal,boyd2011distributed,goldstein2014fast},  Problem (\ref{eq:admm_problem}) can be solved with ADMM iterations. In the $k^{th}$ iteration, it needs to address the following,
\begin{align}  \label{eq:admm_sub_problem_a}
\widehat{ \mathbf W}^{k+1} =  & \argminA_{\widehat{ \mathbf W}} \quad \| \widehat{ \mathbf W} \mathbf X - \mathbf W \mathbf X  \|_2^2   + \frac{\rho}{2} \| \widehat{ \mathbf W} - \mathbf{Z}^k + \mathbf{U}^k  \|_2^2,  
\end{align}
\begin{align} \label{eq:admm_sub_problem_b}
 \mathbf Z^{k+1} = &  \argminA_{\mathbf Z}   \quad  g(\mathbf Z)   + \frac{\rho}{2} \|  \widehat{ \mathbf W}^{k+1}  - \mathbf{Z} + \mathbf{U}^k  \|_2^2,  
 \end{align}
 \begin{align}  
 \mathbf U^{k+1} =  & \mathbf U^k + \mathbf W^{k+1} - \mathbf Z^{k+1}, 
\end{align}
Problem (\ref{eq:admm_problem}) is split into multiple sub-problems with Problem (\ref{eq:admm_sub_problem_a}) and (\ref{eq:admm_sub_problem_b}).  

Problem (\ref{eq:admm_sub_problem_a})  is   similar to Ridge regression problem.  We can directly obtain its solution as   
 \begin{align} 
\widehat{ \mathbf W}^{k+1} = & (\mathbf{X} \mathbf{X}^T + \rho \mathbf{I})^{-1} (\mathbf{X} \mathbf{X}^T \mathbf{ W}^T + \rho (\mathbf Z^k  - \mathbf U^k )^T),  
 \end{align}

To solve Problem (\ref{eq:admm_sub_problem_b}),  we can set $\mathbf Z^{k+1} =  \left( \widehat{ \mathbf W}^{k+1} +  \mathbf U^k \right)  $  and project $\mathbf Z^{k+1} $  on the $g$ function as follows, 
 \begin{align} 
\mathbf Z^{k+1} = & \left( \widehat{ \mathbf W}^{k+1} +  \mathbf U^k \right) \odot \mathbf{M}, 
 \end{align}

 Thus, we can obtain the solution in Theorem 3.1.

\section{SliceGPT Comparison} \label{app:sec:slicegpt_ptb_calibrate}
For SliceGPT~\cite{ashkboos2024slicegpt}, the generated results are sensitive to the calibration datasets, we further show the results of SliceGPT with the calibration of PTB training dataset and 2048 sequence length in Table~\ref{tab:llama_7b_ptb_slicegpt}.
We can know that our method can achieve better performance with the calibration on WikiText2 instead of PTB dataset than the SliceGPT with the calibration on the same dataset.

\begin{table}[]
\centering
\caption{
Compare with SliceGPT using
LLaMA-7B perplexity ($\downarrow$) results on PTB dataset.
}
\label{tab:llama_7b_ptb_slicegpt}
\vspace{-0.2cm}
\resizebox{0.8\linewidth}{!}{
\begin{tabular}{c|c|ccccc}
\toprule
Calibration Dataset & Method & 90\%   & 80\%   & 70\%   & 60\%     & 50\%     \\
\midrule
PTB                 & SliceGPT         & 38.47  & 43.23  & 51.38  & 63.14    & 118.78   \\
WikiText2           & SliceGPT         & 133.80 & 143.89 & 583.58 & 51.86.06 & 15333.66 \\
WikiText2           & Ours             & \textbf{32.05}  & \textbf{36.06}  & \textbf{45.26}  & \textbf{62.07}    & \textbf{117.06}  \\
\bottomrule
\end{tabular}
}
\end{table}

\section{Search on PTB Dataset}
\label{app:sec:search_on_ptb}
To further verify the generalization and effectiveness of our method, we utilize the training portion of the PTB dataset in our search instead of WikiText2. The results, shown in Table~\ref{tab:llama_7b_search_eval_different_dataset}, are evaluated with a sequence length of 2048 using the LLaMA-7B model. Our findings reveal that the models generated through searches on two different training datasets achieve similar performance, demonstrating the robustness and generalization capability of our search method across different datasets.

\begin{table}[t]
\centering
\caption{
LLaMA-7B perplexity ($\downarrow$) results on different datasets of search and evaluation.
}
\vspace{-0.2cm}
\label{tab:llama_7b_search_eval_different_dataset}
\resizebox{0.56\linewidth}{!}{
\begin{tabular}{c|c|ccccc}
\toprule
\multicolumn{2}{c|}{Dataset} & \multicolumn{5}{c}{Inheriting Ratio}   \\
\midrule
Search        & Eval        & 90\%  & 80\%  & 70\%  & 60\%  & 50\%   \\
\midrule
Wiki          & Wiki        & 6.10  & 6.89  & 8.28  & 10.21 & 15.48  \\
Wiki          & PTB         & 32.05 & 36.06 & 45.26 & 62.07 & 117.06 \\
PTB           & Wiki        & 6.22  & 7.08  & 8.41  & 11.08 & 17.23  \\
PTB           & PTB         & 31.24 & 34.87 & 43.89 & 59.07 & 108.83 \\
\bottomrule
\end{tabular}
}
\end{table}

\section{Ablation for 128 Sequence Length }
\label{app:sec:ablation_128_llama13b}

We also present the results for the LLaMA-13B model with a sequence length of 128 in Table~\ref{tab:llama_13b_128_length_results}, demonstrating that our method continues to achieve superior performance.
The results are evaluated with on WikiText2 dataset.

\begin{table}[t]
\centering
\caption{
LLaMA-13B perplexity ($\downarrow$) results on WikiText2 dataset with 128 sequence length.
}
\vspace{-0.2cm}
\label{tab:llama_13b_128_length_results}
\resizebox{0.56\linewidth}{!}{
\begin{tabular}{c|cccc}
\toprule
Ratio & LLM-Pruner(e1) & SliceGPT & FLAP                      & Ours  \\
\midrule
90\%     & 13.23          & 12.68    & 12.25 & \textbf{12.18} \\
80\%     & 16.01          & 15.66    & 13.66                     & \textbf{13.25} \\
70\%     & 21.85          & 21.44    & 15.65                     & \textbf{15.17} \\
60\%     & 31.17          & 32.77    & 18.53 & \textbf{18.14} \\
50\%     & 236.24         & 52.92    & 24.20                     & \textbf{22.65} \\
\bottomrule
\end{tabular}
}
\end{table}

\begin{figure*}[b]
\begin{minipage}{0.485\textwidth}

  \centering
  \includegraphics[width=1.0\textwidth]{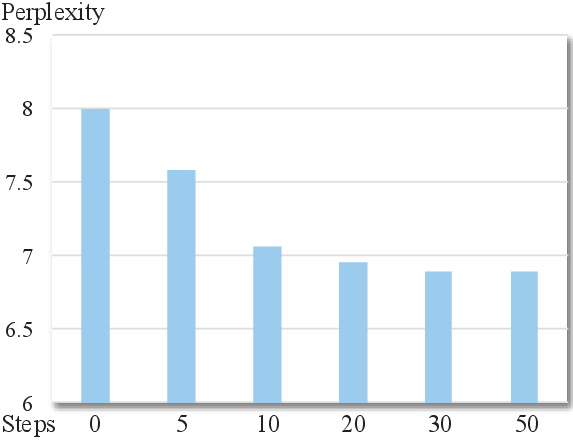}
  \vspace{-0.7cm}
  \caption{
  Ablation analysis for reformation with different numbers of steps.
  }
  \label{fig:ablation_admm_steps}
  \end{minipage}\hfill
\begin{minipage}{0.485\textwidth}
\centering
  \includegraphics[width=1.0\textwidth]{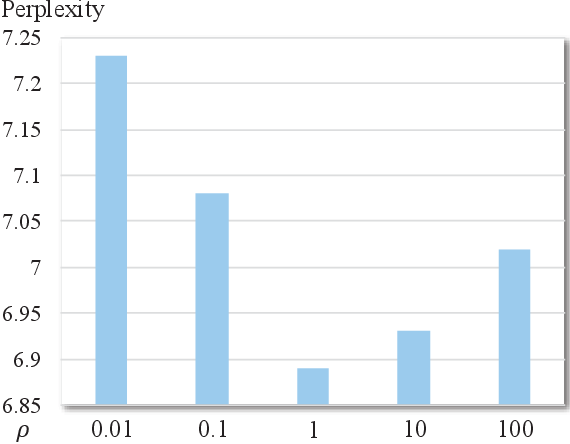}
  \vspace{-0.7cm}
  \caption{
  Ablation analysis for reformation with different $\rho$.
  }
  \label{fig:ablation_rho}
\end{minipage}
\end{figure*}

\section{Ablation in Reformation}
\label{app:sec:ablation_reformation}

To explore the influence of the number of iterations on the reformation process, we conducted experiments with varying iteration counts, as shown in Figure~\ref{fig:ablation_admm_steps}.
The results, evaluated using the LLaMA-7B model with an 80\% inheritance ratio on the WikiText2 dataset with a sequence length of 2048, indicate that model performance improves with an increasing number of iterations, peaking around 30 steps.
Beyond this point, there is minimal difference between 30 and 50 iterations.
Additionally, we examined the impact of different $\rho$ values on the reformation, as depicted in Figure~\ref{fig:ablation_rho}. Our findings show that for $\rho \in [0.01, 0.1]$, the reformed model's performance remains increasing, but deteriorates when $\rho$ reaches 10 or bigger. Hence, $\rho=1$ becomes the optimal value.



\end{document}